\documentclass[11pt]{article}
\usepackage[pdftex]{graphicx}
\usepackage{amsmath}    
\usepackage[square,comma,numbers,sort]{natbib}
\usepackage{url}
\usepackage{authblk} 

\usepackage[font={small}]{caption}

\usepackage{setspace}
\usepackage{multicol}
\title{Automatic time-series phenotyping using massive feature extraction}
\author[1]{B. D. Fulcher\footnote{Lead Contact}}
\author[2]{N. S. Jones}
\affil[1]{Monash Institute of Cognitive and Clinical Neurosciences (MICCN), Monash University, Victoria, Australia}
\affil[2]{Department of Mathematics, Imperial College London, London, United Kingdom}
\date{}

\begin{document}
\maketitle

\paragraph{Summary}
Across a far-reaching diversity of scientific and industrial applications, a general key problem involves relating the structure of time-series data to a meaningful outcome, such as detecting anomalous events from sensor recordings, or diagnosing patients from physiological time-series measurements like heart rate or brain activity.
Currently, researchers must devote considerable effort manually devising, or searching for, properties of their time series that are suitable for the particular analysis problem at hand.
Addressing this non-systematic and time-consuming procedure, here we introduce a new tool, \emph{hctsa}, that selects interpretable and useful properties of time series automatically, by comparing implementations over 7\,700 time-series features drawn from diverse scientific literatures.
Using two exemplar biological applications, we show how \emph{hctsa} allows researchers to leverage decades of time-series research to quantify and understand informative structure in their time-series data.\\

Time-series data, repeated measurements of a quantity taken through time, are being recorded in increasing volumes in biology and medicine.
This wealth of data has opened the door to a range of new research problems, including diagnosis of pathology from biomedical data streams in human patients \cite{Hripcsak:2013hs, Insel:2010cj}, understanding the role of specific neural circuits for behavior \cite{Vogelstein:2014hn}, and linking genotype to phenotype to understand gene function \cite{Nolan:2000kx, Brown:2013ew, Kain:2013hn} or disease processes \cite{Johnson:2006eo, Gates:2011cv, Yang:2014dt}.
While differences in scalar phenotypes are relatively simple to calculate (such as the body length of a worm or the blood pressure of a human subject), it is less clear how to compare complex time-varying data streams (such as the movement dynamics of a worm, the heart rate fluctuations of a clinical patient, or the sequence of reaction times across a cognitive task).
In all of these diverse applications, we require a method for reducing complex time-series data streams to informative, low-dimensional summaries.

A common way of summarizing a time series is by measuring a simple statistic such as its sample mean, which has the advantage of being easily interpretable---e.g., knocking out the gene \emph{unc-9} decreases the mean movement speed of the nematode worm, \emph{Caenorhabditis elegans} \cite{Yemini:2013bd}.
However, this approach fails for many real-world applications in which the phenotypic differences are more subtle than simple mean shifts.
Sophisticated tools for measuring structure in time-series data have been developed by a broad range of researchers, including contributions from the fields of statistics, electrical engineering, economics, statistical physics, dynamical systems, and biomedicine.
This interdisciplinary literature includes summaries of the distribution of values in the data (e.g., Gaussianity, properties of outliers), autocorrelation structure (including power spectral measures), stationarity (how properties change over time), information theoretic measures of entropy and temporal predictability, linear and nonlinear model fits to the data, and methods from the physical nonlinear time-series analysis literature \cite{Fulcher:2013ft}.
There is currently no systematic way of leveraging this giant corpus of scientific work to determine which of these thousands of possible summary statistics best address a particular scientific hypothesis, because the methods have typically been locked in discipline-specific journal articles.
Here we introduce a software package for performing highly comparative time-series analysis, \emph{hctsa}, that opens up the interdisciplinary time-series analysis literature in the form of over 7\,700 \emph{features}, each of which captures a different type of interpretable structure in a univariate time series.
By comparing the performance of these features on a given dataset, \emph{hctsa} facilitates data-driven, statistically-controlled selection of informative time-series summary statistics for phenotyping applications, overcoming an otherwise time-consuming and subjective manual task.

\begin{figure}[h!]
  \centering
    \includegraphics[width=.99\textwidth]{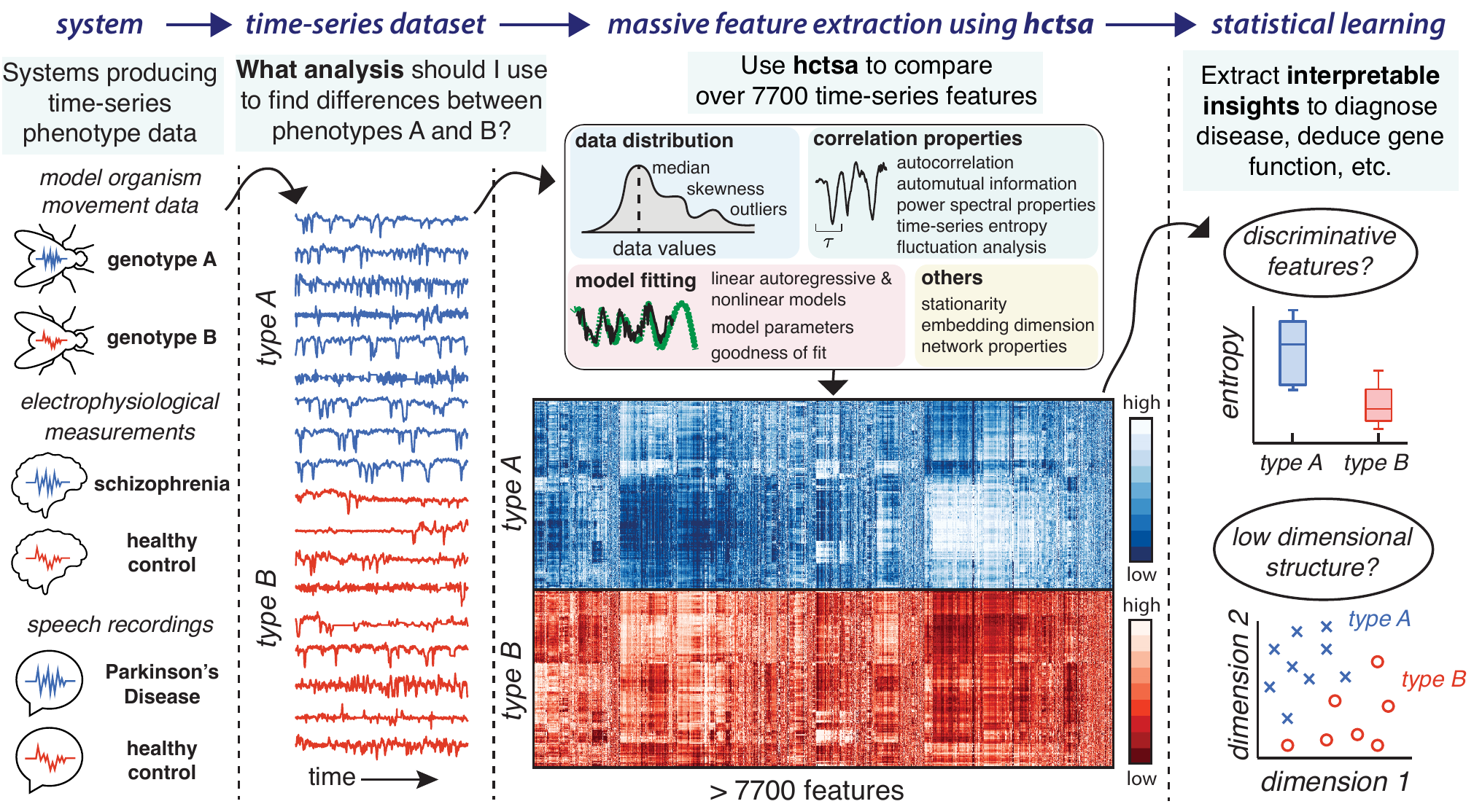}
  \caption{
  \textbf{Using a massive interdisciplinary library of time-series analysis methods to quantify and interpret phenotypic difference using \emph{hctsa}.}
We illustrate the problem of distinguishing two labeled classes of systems using measured time-series data.
The \emph{hctsa} package facilitates massive feature extraction to compare over 7\,700 features of each time series, derived from an interdisciplinary time-series analysis literature.
The feature matrix contains the result of this feature extraction, where each row represents a time series and each column represents a feature that encapsulates some property of that time series (e.g., measures of its autocorrelation structure, entropy, etc.).
Color (blue and red) labels the two types of data---e.g., electrophysiological recordings from healthy controls (A) or people with schizophrenia (B)---and dark/light labels low/high values of each feature, revealing rich structure in the dynamical properties of the dataset.
A range of analysis functions are included with \emph{hctsa}, including learning interpretable differences between the labeled groups (visualized as a box plot revealing that time series of type A have increased entropy), and visualizing informative low-dimensional structure in the dataset.
  }
  \label{fig:overview_schematic}
\end{figure}

The general problem is depicted in Fig.~\ref{fig:overview_schematic}, where we focus on distinguishing time series recorded from two different classes for demonstration (e.g., a patient group and a control group), although we note that the same general framework applies to multiclass classification or regression problems \cite{Fulcher:2013ft}.
After measuring time-series data, \emph{hctsa} is used to perform massive feature extraction, where the behavior of over 7\,700 different scientific analysis methods applied to the dataset can be visualized as a feature matrix with a row for every time series and a column for every feature, shown in Fig.~\ref{fig:overview_schematic}.
The rich structure of the feature matrix reveals some sets of features (i.e., areas of the scientific time-series analysis literature) that capture meaningful differences between groups (e.g., lighter color for type A and darker color for type B in Fig.~\ref{fig:overview_schematic}) and thus represent promising candidates as quantitative phenotypes for distinguishing data of the two types.
In addition to facilitating massive feature extraction, \emph{hctsa} includes a comprehensive suite of analytics for extracting useful and interpretable insights into the structure of the dataset, including:
(i) identifying scientific methods that best quantify differences between labeled groups of data, providing interpretable insights into the phenotypic differences (incorporating permutation testing to statistically control for multiple hypothesis testing),
(ii) building a classifier that draws on the full diversity of scientific methods to optimize the accuracy of phenotypic classification, and
(iii) visualizing low-dimensional structure in the dataset to understand potential clustering structure or other relationships between the time series.
By leveraging a comprehensive interdisciplinary literature on time-series analysis, \emph{hctsa} thus enables researchers to gain a range of interpretable and useful insights into their data.

\begin{figure}
  \centering
    \includegraphics[width=.90\textwidth]{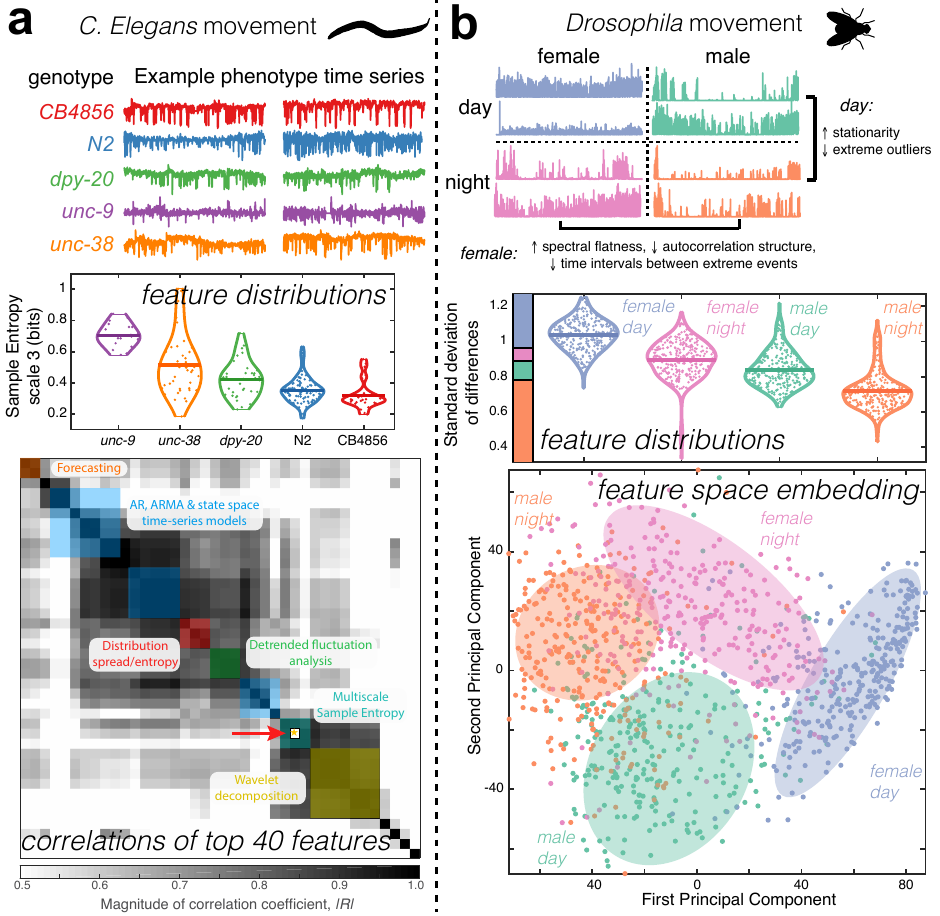}
  \caption{
  \textbf{\emph{hctsa} uncovers interpretable, quantitative phenotypic differences in movement speed time series of \emph{C. elegans} and \emph{Drosophila}.}
  \textbf{A} \emph{C. elegans}. Top: Two examples of movement speed time series are shown for each of five genotypes.
  Middle: Class distributions of multiscale Sample Entropy, selected by \emph{hctsa} as an informative measure, are shown as a violin plot, demonstrating that the neural mutant \emph{unc-9} genotype has the highest average Sample Entropy at this scale, followed by \emph{unc-38}, the morphological mutant \emph{dpy-20}, the lab-based strain N2, and the wild type strain CB4856.
    Bottom: The top 40 features identified by \emph{hctsa} for distinguishing the five genotypes span a wide range of time-series analysis techniques, labeled using color, and form sets of highly correlated groups.
    The multiscale Sample Entropy shown above is indicated with a red arrow and star.
  \textbf{B} \emph{Drosophila}.
  Top: Two examples of movement speed time series are shown for each of four groups, labeled as either `male' or `female', and either `day' or `night'.
    Interpretable measures of difference between each pair of conditions were extracted using \emph{hctsa}, and are summarized using text.
  Middle: \emph{hctsa} identified the standard deviation of successive changes in movement speed as a simple but highly discriminative feature, shown as a violin plot.
  Bottom: A two-dimensional principal components projection of the dataset across the full \emph{hctsa} feature library is informative of the class structure in the dataset.
  Shading has been added to guide the eye.
  }
  \label{fig:phenotyping}
\end{figure}

To demonstrate the approach, we applied \emph{hctsa} to two case studies of \emph{C. elegans} and \emph{Drosophila melanogaster} movement, as shown in Fig.~\ref{fig:phenotyping}.
Sample time series of the movement speed dynamics of five different strains of \emph{C. elegans} \cite{Brown:2013ew} are shown in Fig.~\ref{fig:phenotyping}a (upper).
Being noisy empirical recordings with no clear visual differences between strains, it is unclear what types of analysis methods might capture differences between the genotypes.
Using \emph{hctsa} to compute the behavior of over 7\,700 different time-series features (subsequently filtered down to 6\,504 well-behaved features, see \emph{Online Methods}), we found that the feature set as a whole predicted genotype from the short, noisy time series with a ten-fold cross-validated balanced accuracy of 80\% (using a linear SVM; chance level: 20\%).
A total of 4\,499 different features of movement speed time series were individually informative of the genotype label ($q < 0.05$, FDR-corrected, permutation test), with the 40 most informative features spanning diverse methodological literatures, including AR and state space model fitting methods, detrended fluctuation analysis, local mean forecasting, multiscale Sample Entropy, and wavelet decompositions of the signal, shown as a structured pairwise correlation plot in Fig.~\ref{fig:phenotyping}a (lower).
One of the multiscale entropy measures \cite{Costa:2005hi} is highlighted in Fig.~\ref{fig:phenotyping}a (middle), which computes the Sample Entropy, $\text{SampEn}(2,0.15)$, at a scale level 3 (corresponding to 100\,ms bins), which can be thought of as quantifying the `unpredictability' of the time series at this timescale.
Violin plots in Fig.~\ref{fig:phenotyping}a reveal physiologically interpretable differences between the genotypes, with the lab strain N2 and wild isolate strain CB4856 showing the most predictable time series at this timescale, the morphological mutant, \emph{dpy-20} being intermediate, and neural mutants like the \emph{unc-38} and \emph{unc-9} being the least predictable.
Along with many other time-series properties discovered by \emph{hctsa}, we thus find that this multiscale entropy measure constitutes a useful quantitative phenotype for \emph{C. Elegans} movement, that also provides new interpretable insights beyond simple summaries like the mean speed.
For example, the two neural mutants \emph{unc-38} (which encodes a nicotinic acetylcholine receptor alpha subunit) and \emph{unc-9} (which encodes a structural component of gap junctions) display similar movement speeds, but our analysis shows that they affect movement distinctly, exhibiting significant differences in their movement predictability.
Furthermore, the selection of this multiscale entropy measure, achieved automatically using \emph{hctsa}, mirrors detailed manual research proposing the similar concept of `compressibility' of posture sequences as a quantitative phenotype for \emph{C. elegans} \cite{GomezMarin:2015jt}.
By comparing a wide interdisciplinary literature of analysis methods, \emph{hctsa} thus selects biologically informative quantitative phenotypes for these time-series data.

We also applied \emph{hctsa} to 12\,h movement speed time series of \emph{Drosophila} restricted to a one-dimensional tube, labeled as either `day' (light on) or `night' (light off), and as either `male' or `female', as shown in Fig.~\ref{fig:phenotyping}b \cite{Gilestro:2012en}.
Leveraging the full feature library, we successfully classified recordings made during the day versus at night with a mean 10-fold cross-validated balanced accuracy of 98\%, and also clearly distinguished the sex of the organisms (96\%).
The \emph{hctsa} package selected interpretable quantitative time-series features for different groupings of the data, as annotated in Fig.~\ref{fig:phenotyping}b (upper), highlighting the increased spectral flatness and shorter durations between outliers in females relative to males (driven by their less predictable movement), and increased stationarity of movement dynamics during the day, with fewer extreme outliers than at night (which is dominated by more bursty patterns between sleep and activity).
Taking four combination classes (colored in Fig.~\ref{fig:phenotyping}b), the mean balanced 10-fold cross-validated balanced accuracy remains high, at 95\%.
Again, \emph{hctsa} identifies interpretable features like the standard deviation of incremental differences in the $z$-scored time series, shown as a violin plot in Fig.~\ref{fig:phenotyping}b (middle).
This elementary measure gives higher values to time series exhibiting greater changes from one time point to the next, providing a simple but easily interpretable measure of temporal predictability, which is increased during the day, and in females.
Even when class labels are not used, \emph{hctsa} can draw on the combined behavior of thousands of scientific methods to provide an informative low dimensional principal components representation of the dataset, shown in Fig.~\ref{fig:phenotyping}b (lower), in which the four classes are clearly separated.
The subtle quantitative phenotypes of \emph{Drosophila} movement provided automatically by \emph{hctsa} go beyond simple comparisons of the overall amount of movement between day and night, or between males and females \cite{Gilestro:2012en, Isaac:2010jv}.
For example, while it is known that females have shorter sleep bouts than males, \emph{hctsa} quantifies the sexually dimorphic behavior by selecting new measures of, for example, predictability of movement (reduced in females) or time intervals between large movements (reduced in females).
The use of \emph{hctsa} thus provides us with a picture of more erratic female \emph{Drosophila} movement that may reflect their need to forage for food and select egg laying sites \cite{Isaac:2010jv}, in contrast to the more predictable male behavior of conserving energy to avoid predators.
Our results demonstrate the benefits of leveraging a wide variety of time-series analysis methods to automatically learn and interpret structure in biological time-series data.
In summary, we introduce a new software framework, \emph{hctsa}, which automates the selection of quantitative phenotypes from time-series data by leveraging a large and interdisciplinary literature on time-series analysis.
In a reversal of the typical time-series analysis process in which methods are selected manually by researchers, here we show that statistical machine learning can aid this process of human learning by subjecting time-series data to thousands of scientific methods.
In addition to the high throughput phenotyping applications demonstrated here, \emph{hctsa} has general utility, including behavioral phenotyping in cognitive science, and diagnosis of disease from biomedical data streams such as heart rates or brain dynamics.
Furthermore, although we focus on classification problems here, we note that the same approach applies to regression problems, where one aims to find time-series features that vary with a continuous variable (such as the dosage of a drug, a standardized depression score of a patient, etc.) \cite{Fulcher:2013ft}.
Code for running \emph{hctsa} in Matlab has been developed and refined over many years through a range of diverse applications \cite{Fulcher:2012gj, Fulcher:2013ft, Fulcher:2014uo} and is available at \url{www.github.com/benfulcher/hctsa}, with accompanying comprehensive documentation at \url{www.gitbook.com/book/benfulcher/hctsa-manual}.

\paragraph{Acknowledgements}
We thank Andre Brown and Bertalan Gyenes for sharing the \emph{C. elegans} movement dataset, and helpful feedback on the resulting analysis and manuscript.
We thank Giorgio Gilestro and Quentin Geissmann for sharing the \emph{Drosophila} movement dataset, and helpful feedback on the resulting analysis.
Many thanks to Rachael Fulcher for help with graphic design, and to Alex Fornito and Iain Johnston for useful feedback on the manuscript.


\clearpage
\setcounter{page}{1}
\section*{Supplementary Information}


\subsection*{Software details and reproducibility}
Following from the original concept and proof of principle for a highly comparative approach to time-series analysis \cite{Fulcher:2013ft}, this article introduces a well-documented and user-friendly Matlab-based software platform for performing it (Matlab is a product of The MathWorks, Natick, MA).
The set of over 7\,700 features has been developed and refined through applications to a wide range of research and industrial problems over many years.
A full analysis pipeline has also been built to allow researchers to run highly comparative analysis on their own data, including functions for initiating new analysis tasks, computing features locally in Matlab or through an interface to a mySQL server (enabling distributed computing for large datasets), processing the results of the feature extraction (including options for filtering features on their behavior and feature normalization), and a range of other analytic outputs to facilitate scientific interpretation (including the plots shown in this paper).

\subsubsection*{Dataset availability}
The two datasets analysed here, including the labeled time series and the full results of \emph{hctsa} feature extraction, are available in the form of Matlab files (.mat) for \emph{C. elegans}: \url{https://dx.doi.org/10.4225/03/580478f951263}, and for \emph{Drosophila}: \url{https://dx.doi.org/10.4225/03/5804798d2a2ec}.

\subsubsection*{Code availability}
Analyses presented here were computed using v0.92 of \emph{hctsa}, which contains a total of 7\,749 features.
The \emph{hctsa} software is freely available at \url{github.com/benfulcher/hctsa/}.
Analysis pipelines used to produce the results reported here (as well as many other outputs from \emph{hctsa}) are available at \url{github.com/benfulcher/hctsa_phenotypingWorm/} and \url{github.com/benfulcher/hctsa_phenotypingFly/} for the \emph{C. elegans} and \emph{Drosophila} datasets, respectively.

\subsection*{Analysis details}

\subsubsection*{Feature filtration and normalization}
For any given analysis, we filtered out any features that were constant across the dataset or contained any `special' values (e.g., due to applying a method that is inappropriate for the data, such as fitting a positive-only distribution to data that are not positive only, or attempting to fit a model to the data that does not converge, etc.).
Due to this filtering, a different number of total features will be usable for a given dataset, depending on its properties.

When searching for discriminative individual features, we did not normalize or rescale feature values to enable results to be interpreted in the natural scale of each feature.
However, when computing the Principal Components of a dataset, or learning a classifier in the full feature space, we normalized each feature to the unit interval using a scaled robust sigmoid function \cite{Fulcher:2013ft}:
\begin{equation}\label{eqn:scaledSQzscore}
\hat{\mathbf{f}} = \left[1+\exp\left(-\frac{\mathbf{f}-m_\mathbf{f}}{1.35 r_\mathbf{f}}\right)\right]^{-1},
\end{equation}
where $\hat{\mathbf{f}}$ represents the normalized feature values across a time-series dataset, $\mathbf{f}$ is the vector of un-normalized feature values, $m_\mathbf{f}$ is the median of $\mathbf{f}$, and $r_\mathbf{f}$ is its interquartile range.


\subsubsection*{Classification}
For multi-class classification, we trained linear support vector machine classifiers in Matlab 2015b (a product of The MathWorks, Natick, MA) using the \texttt{fitcecoc} function with a linear kernel SVM.
To compare single univariate features, we used simple linear discriminant analysis (using \texttt{classify}).
When training SVM classifiers, we weighted each observation, $x$, as the inverse probability of its class label across the dataset to account for class imbalance.

Due to imbalance of observations across the multiclass classification problems investigated here, we report balanced classification accuracy, $C_\mathrm{bal}$, over $m$ classes in terms of the number of correctly identified examples of a given class $t_i$, and the total number of examples of each class, $c_i$, as
\begin{equation}
    C_\mathrm{bal} = \frac{1}{m} \sum_{i=1}^m t_i/c_i .
\end{equation}
Balancing the accuracy in this way ensures that all classes contribute equally to the classification statistic.

\subsection*{Datasets}
\subsubsection*{\emph{Caenorhabditis elegans} movement speed}
Movement speed time-series data were obtained from approximately 15\,min videos obtained using tracking microscopes \cite{Yemini:2013bd, Brown:2013ew} (see \url{wormbase.org} for more information).
A total of 226 movement time series sampled at 30.03\,Hz were obtained from the CB4856 (Hawaiian wild isolate, 29 time series) and N2 (lab strain, 100 time series) strains, and the mutants \emph{dpy-20} (34 time series), \emph{unc-9} (20 time series), \emph{unc-38} (43 time series).
For the \emph{dpy-20}, \emph{unc-9}, \emph{unc-38} knockouts and the CB4856 strain, all available data at the specified frame rate were used.
For the wildtype N2 strain, we took a random sample of 100 of the 1\,200 time series recorded at a sampling rate of 30.03\,Hz.
If missing data in a time series made up less than 15\% of its length and in a contiguous block at the beginning or end of the recording, the time series was retained with this section of missing data removed, otherwise the time series was removed.

\subsubsection*{\emph{Drosophila melanogaster} movement speed}
We analysed time series of the movement speed of flies restricted to a one-dimensional tube and tracked continuously for between 3 and 6 days using video tracking \cite{Gilestro:2012en, Donelson:2012bz}.
Movement speed was estimated as the maximum speed of the measured data (sampled at approximately 2\,Hz) in each non-overlapping 10\,s time window, where displacements are measured as the euclidean distance between successive co-ordinates of the fly.
In this way, here we analyze these time series of movement speed, sampled at a rate of 0.1\,Hz.
Time series were split into 12\,h segments and labeled as either `day' (light on, 574 time series) or `night' (light off, 574 time series), and as either `male' (554 time series) or `female' (594 time series).



\end{document}